\documentclass{ecai} 

\usepackage{latexsym}
\usepackage{amssymb}
\usepackage{amsmath}
\usepackage{amsthm}
\usepackage{booktabs}
\usepackage{enumitem}
\usepackage{graphicx}
\usepackage{color}
\usepackage{array}
\usepackage{tabularx}

\newcommand{\BibTeX}{B\kern-.05em{\sc i\kern-.025em b}\kern-.08em\TeX}

\begin{document}


\begin{frontmatter}

\paperid{1} 

\title{Knowledge-Augmented Reasoning for EUAIA Compliance and Adversarial Robustness of LLMs}

\author[A]{\fnms{Tomas}~\snm{Bueno Momcilovic}\thanks{Corresponding Author. Email: momcilovic@fortiss.org \\\textbf{Accepted Manuscript.}}}
\author[A]{\fnms{Dian}~\snm{Balta}}
\author[B]{\fnms{Beat}~\snm{Buesser}}
\author[C]{\fnms{Giulio}~\snm{Zizzo}}
\author[C]{\fnms{Mark}~\snm{Purcell}}

\address[A]{fortiss Research Institute of the Free State of Bavaria, Munich, Germany}
\address[B]{IBM Research Europe, Zurich, Switzerland}
\address[C]{IBM Research Europe, Dublin, Ireland}

\begin{abstract}
The EU AI Act (EUAIA) introduces requirements for AI systems which intersect with the processes required to establish adversarial robustness. However, given the ambiguous language of regulation and the dynamicity of adversarial attacks, developers of systems with highly complex models such as LLMs may find their effort to be duplicated without the assurance of having achieved either compliance or robustness. This paper presents a functional architecture that focuses on bridging the two properties, by introducing components with clear reference to their source. Taking the detection layer recommended by the literature, and the reporting layer required by the law, we aim to support developers and auditors with a reasoning layer based on knowledge augmentation (rules, assurance cases, contextual mappings). Our findings demonstrate a novel direction for ensuring LLMs deployed in the EU are both compliant and adversarially robust, which underpin trustworthiness.
\end{abstract}

\end{frontmatter}


\section{Introduction}

The European Union (EU) bases trustworthiness of artificial intelligence systems (AIS) on three properties: lawful, ethical, and robust \cite{eu2019trustworthy}. The EU AI Act (EUAIA, \cite{euaia2024corrigendum}) is an upcoming regulation that sets obligations on the lawful design and implementation of AIS in the EU. Its content outlines the high-level requirements for improving the auditability of the AIS, whose generic descriptions are interepretable across contexts. 

However, for properties such as adversarial robustness, providers of AIS with large language model-based (LLM) components are facing a difficult and highly dynamic challenge whose boundaries are not yet known. Providers in the EU who would like to ensure both compliance and robustness, are doubly burdened. First there is the need to constantly readapt their defenses against novel adversarial attacks \cite{geiping2024}, and second is the overhead for correctly interpreting "compliant robustness" with auditable evidence.

This paper presents a novel approach of knowledge augmentation for aligning adversarial robustness of LLMs with EUAIA compliance. By integrating detection, reasoning and reporting layers alongside the layer for interacting with users, we propose a comprehensive functional architecture as a reference for ensuring the AIS is dynamically protected and auditable. The research provides a framework for combining robustness and compliance activities while retaining the provenance to the requirements.

Our roadmap centers on solution-oriented requirements engineering \cite{pohl2010requirements} and knowledge augmentation (i.e., knowledge representation and reasoning \cite{keet2023}) to develop the architecture of our prototype. This process of creating a blueprint of a compliant LLM defense against adversarial attacks involves three steps.

First, we extract the legal duties and relevant stakeholders from the EUAIA (\cite{hohfeld2001,vanengers2015law}; cf. \cite{bueno2024assuring} for the expanded list), and structure them into draft requirements in the next section. This approach takes inspiration from \cite{floridi2022capAI}. Second, concepts and relations surrounding LLMs are represented in a simple ontology \cite{ontotext2023b}. State-of-the-art attacks and defenses in the context of natural language tasks are recovered from preprints \cite{Zou2023_Universal, Zeng2024_How, geiping2024}. The third step is a representation of the knowledge in a cyclical process model of actions between stakeholders and components, and the corresponding sources.

\section{Requirements}

\begin{table*}[ht]
    \centering
    \caption{Requirements related to adversarial robustness and their sources}
    \label{tab:requirements}
    \setlength{\extrarowheight}{2pt}
    \begin{tabularx}{\linewidth}{|r|>{\raggedright\arraybackslash}X|>{\arraybackslash}p{0.22\linewidth}|}
        \toprule
        \textbf{id} & \textbf{Requirement} & \textbf{Source} \\
        \midrule
        R0  & Include the following stakeholders: user; (malicious) third party; GPAI provider; AIS provider; \newline GPAI or AIS deployer; national competent authority; market surveillance authority; AI office. & Art. 3 \& Rec. 76 \cite{euaia2024corrigendum}; cf. \cite{bueno2024assuring} \\
        R1  & Include the following roles: user; developer (i.e., system or LLM engineer, researcher, scientist); auditor. & \cite{suresh2021beyond} \\
        R2  & Identify, evaluate and mitigate \textit{reasonably foreseeable} risks of the system. & Art. 9 Para. 2 \cite{euaia2024corrigendum} \\
        R3  & Ensure appropriate and adequate risk management measures. & Art. 9 Para. 5 \cite{euaia2024corrigendum} \\
        R4 & Detect automated attacks such as prompts with randomized perturbations. & \cite{Zou2023_Universal} \\
        R5 & Detect semi-automated attacks such as heuristic-based exploitation of the undertrained aspects of the model. & \cite{geiping2024} \\
        R6  & Establish cybersecurity measures against adversarial and poisoning attacks. & Art. 15 Para. 5 \cite{euaia2024corrigendum} \\
        R7 & Achieve sustained coverage of detected and prevented attacks above a predefined threshold. & \cite{alon2023detecting} \\
        R8  & Establish an appropriate level of robustness and cybersecurity. & Art. 15 Para. 1 \cite{euaia2024corrigendum} \\
        R9 & Provide information about robustness and cybersecurity (e.g., metrics) and their limitations in \newline instructions for use. & Art. 13 Para. 3 \& Annex IV \cite{euaia2024corrigendum} \\
        R10  & Design system for effective human oversight regarding safety monitoring and prevention/minimization of reasonably foreseeable misuse. & Art. 14 Para. 2 \cite{euaia2024corrigendum} \\
        R11  & Design appropriate functionalities for human overseers to monitor for "anomalies, dysfunctions and \newline unexpected performance." & Art. 14 Para. 4 \cite{euaia2024corrigendum} \\
        R12  & Report on measures and tests used for adversarial testing, model alignment, and fine-tuning. & Art. 53 Para. 1 \& Annex XI \cite{euaia2024corrigendum}; Art. 11 \& Annex IV \cite{euaia2024corrigendum} \\
        R13 & Supply information on testing, safeguards and risk mitigation measures at the request of the AI Office. & Art. 92 Para. 5 \& 7 \cite{euaia2024corrigendum} \\
        R14 & Establish and report on the definite, reasonably likely or suspected causal link between the system and \newline a serious incident. & Art. 73 Para. 2-6 \cite{euaia2024corrigendum} \\
        R15 & Detect manual attacks based on patterns of persuasion (i.e., "jailbreaking"). & \cite{Zeng2024_How} \\
        R16  & Notify supervising stakeholder of a serious incident. & Art. 73 Para. 1, 7-8 \& 11 \cite{euaia2024corrigendum} \\
        \bottomrule
    \end{tabularx}
\end{table*}

The EUAIA places AI systems that are deployed in particular products or domains under categories of risk, where high-risk AI systems play a central role \cite{euaia2024corrigendum}. The regulation which has been adopted in 2024 places duties on stakeholders at design and runtime. Before standards are expected in the following years, these duties provide a basis for safety- and security-oriented requirements.

General-purpose AI models (GPAI, Art. 3, \cite{euaia2024corrigendum}) such as LLMs are not inherently high-risk. However, their broad capabilities and wide attack surface have over time crystallized similar requirements with respect to adversarial robustness. In examples provided by an ever-increasing body of work (cf. \cite{Zou2023_Universal}), adversaries of an LLM can include third parties with malicious intentions, curious users who test the boundaries, and even completely benign users whose prompts elicit harmful or otherwise unintended output. 

Based on an analysis of requirements in Table \ref{tab:requirements}, EUAIA compliance and adversarial robustness are complementary properties, despite the difference in details. On the one hand, the requirements that are derived from the regulation (cf. \cite{bueno2024assuring} for expanded list) provide a generic description of stakeholders (R0), risk management (R3) and cybersecurity measures, and the need for human oversight (R10) and reporting (R12). On the other hand, the state-of-the-art literature introduces specific roles (R1), the detection of automated (R4), semi-automated (R5), and manual attacks (R15), and sustained coverage of these threats (R7). However, aside from the direct references to the term in EUAIA (R6, R12), the two sources emphasize different facets of a larger system - i.e., components for assuring the quality of adversarial robustness beyond the purely functional components of an LLM-supported application. In the next section, we introduce one approach to satisfying both facets.

\section{Functional Architecture and Workflow}

\subsection{Architecture}

\begin{figure*}[ht]
    \centering
    \includegraphics[width=0.8\linewidth]{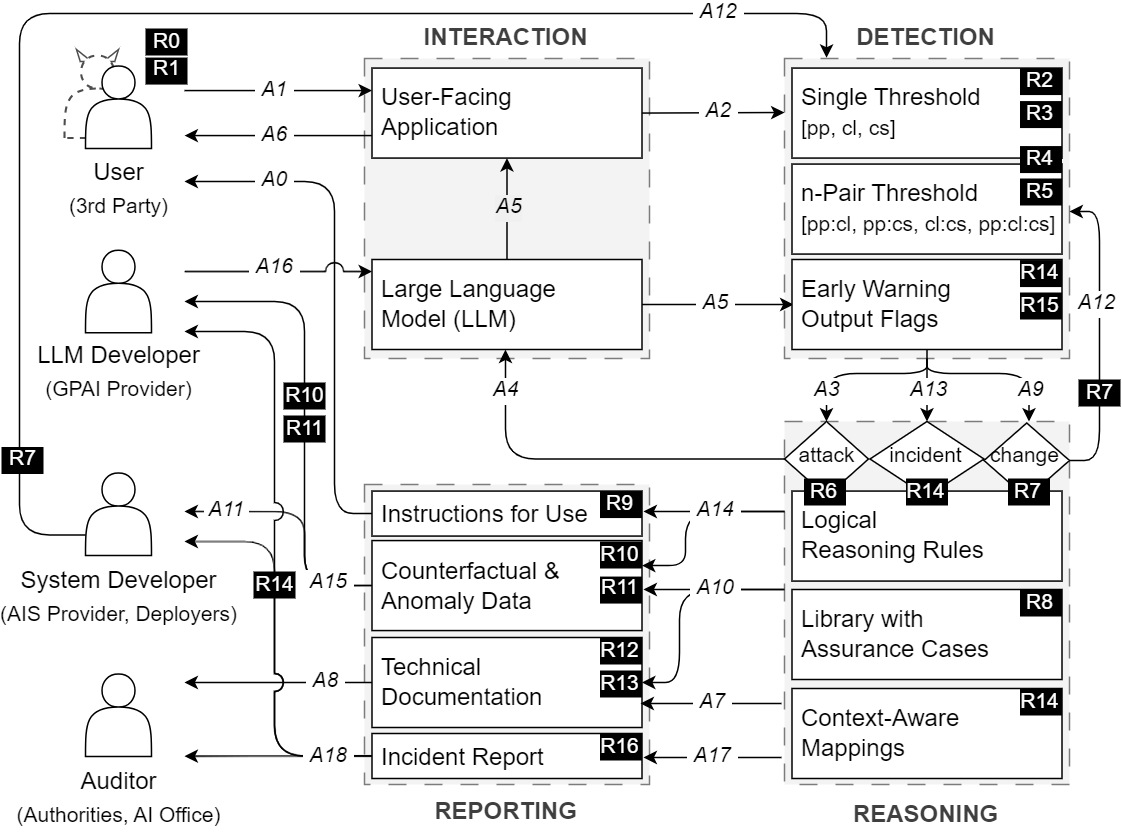}
    \caption{Reference functional architecture for compliant adversarial robustness of LLM-based AI systems.}
    \label{fig:architecture}
\end{figure*}

The functional architecture depicted in Figure \ref{fig:architecture} is composed of a cyclical workflow linking four stakeholders and four layers of components. Stakeholders and components are connected with arrows denoting action IDs (A\#), as described in Table \ref{tab:actions}, whereby each non-functional element of the architecture has a corresponding requirement ID (R\#) identified in Table \ref{tab:requirements}. 

The stakeholders include users, LLM developers, AIS developers and auditors, who represent the various roles involved in the design and implementation of AIS, with the corresponding EUAIA-defined role in parentheses. Auditors and users are external temporary roles, whereby a user can be benign, curious or malicious. Developers are internal and lasting roles, whose responsibilities depend on the access to the internal workings of an LLM and the system deploying it.

The layers involve the interaction layer which fulfils the functional requirements of an AIS, and detection, reasoning and reporting layers which fulfill the quality requirements underlying robustness. In other words, the first layer is enough to establish a fully working AIS, without special consideration for other properties. Interaction has a simple structure inspired by practice \cite{openai2022}, containing the user-facing application (i.e., the interface between the user and the AIS) and the LLM. 

\subsection{Workflow}

\begin{table}[ht]
    \centering
    \caption{Actions and their descriptions.}
    \label{tab:actions}
    \setlength{\extrarowheight}{2pt}
    \begin{tabularx}{\linewidth}{|r|X|}
        \toprule
        \textbf{id} & \textbf{Action} \\
        \midrule
        A0 & Displays relevant information about the LLM, disclaimers, and limitations. \\
        A1 & Enters a prompt. \\
        A2 & Forwards the prompt and the metadata. \\
        A3 & Provides the first batch of classification using the deployed detectors. \\
        A4 & Provides the evaluation with the prompt (if benign) or warning (if malicious). \\
        A5 & Provides the generated result according to the evaluation. \\
        A6 & Displays the generated result. \\
        A7 & Provides data on the metrics and thresholds used for detectors. \\
        A8 & Displays the metrics and relevant data. \\
        A9 & Provides the second batch of classification using all relevant detectors and their combinations. \\
        A10 & Provides a counterfactual assessment comparing the coverage and accuracy of deployed and non-deployed detector combinations. \\
        A11 & Displays the counterfactual assessment. \\
        A12 & Reconfigures the detector combinations and their threshold values. \\
        A13 & Provides flagged LLM output (i.e., anomaly or incident) and the corresponding input prompt. \\
        A14 & Provides the data on the detected anomalies. \\
        A15 & Displays information about the individual or group of anomalies. \\
        A16 & Makes adjustments to the LLM based on the provided data. \\
        A17 & Provides data about the anomalies flagged as incidents. \\
        A18 & Displays information about the (serious) incidents. \\
        \bottomrule
    \end{tabularx}
\end{table}

Detection is based on input and output classification, following the current paradigm of dealing with adversarial attacks \cite{alon2023detecting}. Input detectors have thresholds based on some combination of single and n-pairs of metrics. Metrics denote ways of measuring particular properties of input prompts, examples including perplexity (pp; i.e., the extent to which the model is "surprised" by a prompt), context length (cl) and character set size (cs). Output detectors attempt to detect unexpected LLM results which may be results of undetected attacks. They can be implemented similar as for inputs, but also using flags for harmful keywords to provide an early warning to the developer.

Reasoning serves as the middleware between other layers by decoupling the logic from detection, interaction and reporting activities. The layer provides a set of rules derived using deductive or inductive reasoning, which are behind decisions to classify an input as an attack, an output as an incident, or detector performance as a trigger for change. The library with assurance cases is a set of graphs connecting claims about satisfied requirements relating to compliance and robustness, with the evidence from chosen strategies. Given the adaptability of LLMs and the context-specific properties, context-aware mappings provide the needed metadata to separate the rules and assurance case elements to what they are appropriate. In addition, these mappings enable the variables in reports to be linked with actual values.

Finally, the reporting layer is primarily based on the EUAIA need for human oversight. Instructions for use and technical documentation are factsheets for users and auditors respectively. However, given the relevance of figures and test results to the monitoring of adversarial robustness, these components are useful to developers for AIS debugging and improvement as well. In addition, assessments based on counterfactuals and anomaly data allow the developers to monitor detectors with respect to needed changes. Incident reports are triggered by an event of a potentially successful attack; although primarily an EUAIA requirement for mandatory auditing of serious incidents, less critical but problematic anomalies provide an opportunity to developers to perform forensic analyses.

\begin{figure}[ht]
    \centering
    \includegraphics[width=\linewidth]{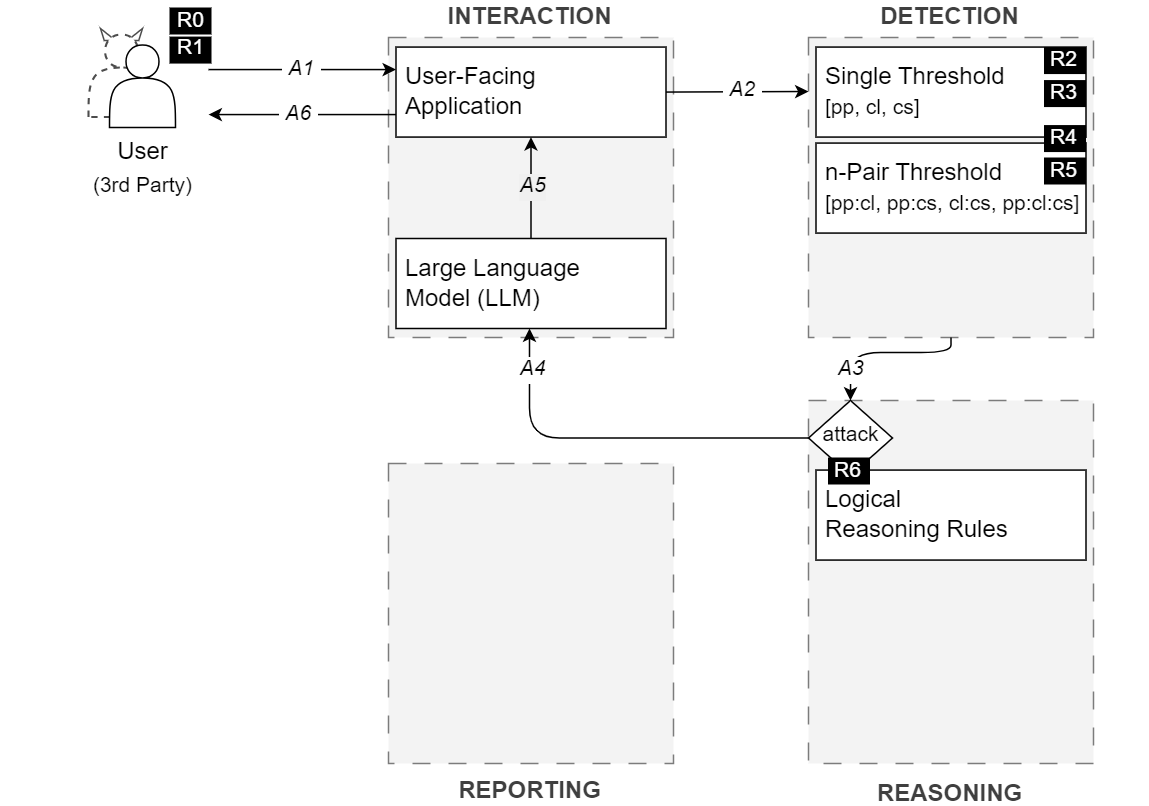}
    \caption{Primary cycle with interaction and basic attack detection.}
    \label{fig:sequential}
\end{figure}

The Figure \ref{fig:sequential} depicts three main cyclic processes. The primary cycle is the simplest: a user enters a prompt into the application (A1), which is then forwarded to the input detectors (A2). The detectors' results are provided as input (A3) to a rule that classifies the prompt as safe or unsafe, passing on the prompt or the warning respectively to the LLM (A4). The LLM then generates  instead elicits a warning to the user (A4, A5, A6). Relying only on this cycle would be a naive approach to handling adversarial attacks, whereby the developer would expect the detectors to perform well over time and prompts.

\newpage
The secondary cycle introduces the required auditability for EUAIA compliance. The information about the deployed detectors is structured in assurance cases, which feed into the documentation (A7). This documentation provides an interface to the user to understand the model and its limitations before use (A0), and an interface to the auditor (A8) to establish a clear picture about the AIS. 

This cycle also provides the basis for required dynamicity for adversarial robustness. Assurance cases are intended to provide the logic needed to evaluate detector performance. Given a number of prompts or some other triggering rule (A9), prompts would be processed through non-deployed detector combinations. This would provide the basis for counterfactually assessing the sustained robustness of the detectors (A10). This evaluation is initially be the responsibility of the AIS developer (A11), whose understanding of the context-sensitive performance and coverage would be needed to reconfigure the detectors (A12).

\begin{figure}[ht]
    \centering
    \includegraphics[width=\linewidth]{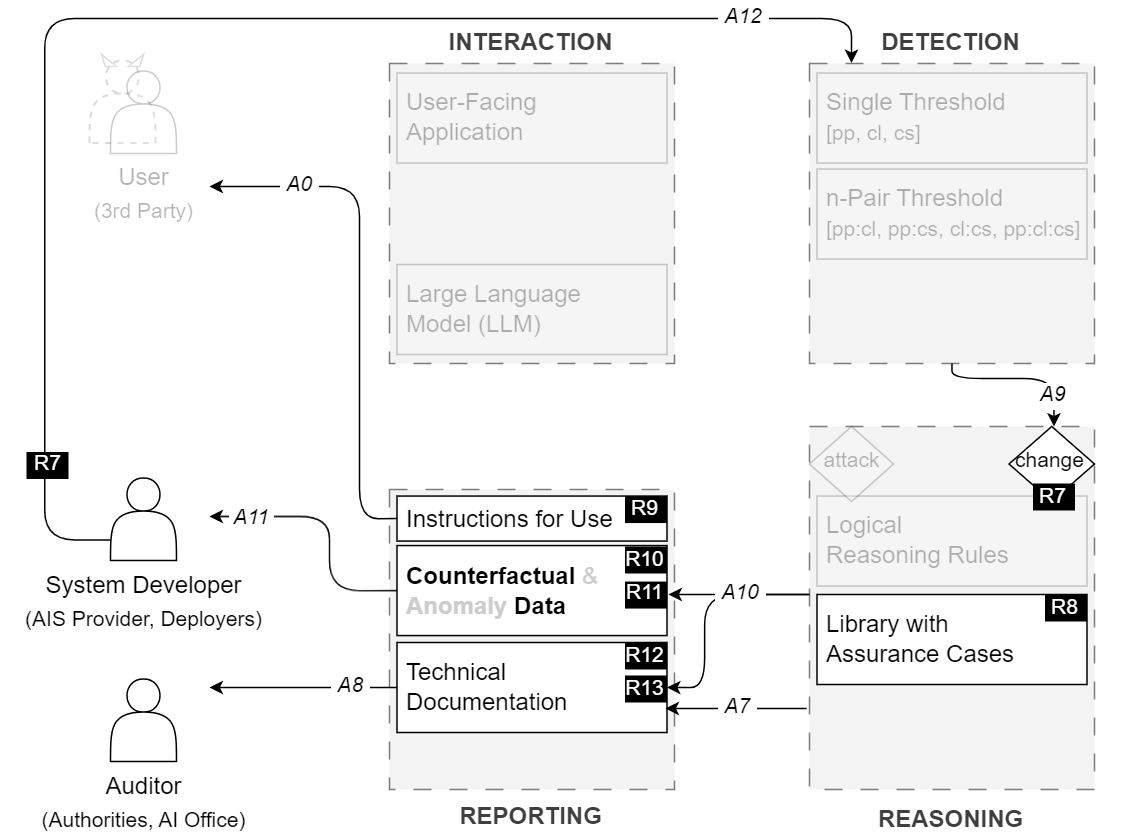}
    \vspace{-0.5cm}
    \caption{Secondary cycle with assurance, monitoring and reporting.}
\end{figure}

The tertiary cycle introduces mechanisms for handling failure systematically. An evaluation of the LLM output (A5), whether in real-time or delayed intervals, allows some successful attacks (i.e., incidents; A13) to be automatically detected. Here is context-specific information necessary to operationalize ambiguous EUAIA language: which risks or anomalies are "reasonably foreseeable" (A14) and worth exploring; which incidents are "serious" enough (A17) to demand contact with the auditor (A18); and when is a given risk management procedure not "suitable" anymore (A12). This cycle also proposes providing relevant information to the LLM developer, who may not be associated with the AIS directly, but nonetheless benefits from adversarially retraining the LLM, thereby making it more secure in the AIS as well.

\begin{figure}[ht]
    \centering
    \includegraphics[width=\linewidth]{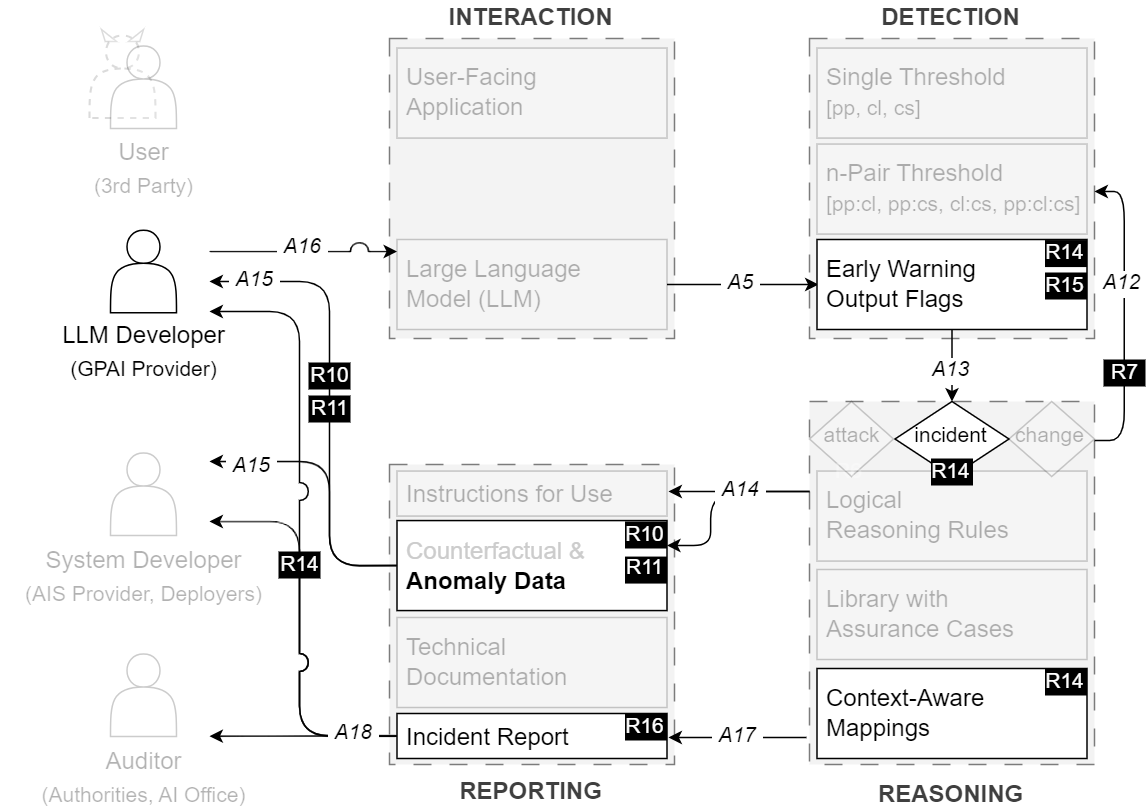}
    \vspace{-0.5cm}
    \caption{Tertiary cycle for advanced handling of failures.}
\end{figure}

\section{Discussion and Conclusion}

This paper introduces a knowledge-augmented framework designed to align the adversarial robustness of large language models with the EU AI Act compliance. By using a combination of detection, reasoning and reporting layers, we address the critical need for compliance and robustness in AI systems. 

The functional architecture is meant as a reference for implementing physical components. For example, our early prototype implements simple detectors in Python, including n-pair detectors based on logistic regression classifiers pretrained on Hugging Face jailbreak data \cite{jaramillo2023jailbreak}. The reasoner is based on a combination of an assurance case and an ontology, stored in the graph database, where evaluations are performed through queries. Additionally, graphical visualizations and textual data is generated in Jupyter Notebooks to provide clear and informative reporting. The interaction layer uses the streamlit package to provide a user-facing application, while GPT-2, accessed via the Hugging Face package, serves as the foundational LLM.

Our findings highlight a promising direction for developing resilient AI technologies capable of withstanding adversarial attacks while meeting regulatory standards. Future work will focus on the following aspects: (1) defining new detectors and combinations thereof, such as classifiers trained on larger samples of malicious and benign prompts; (2) expanding the reasoning based on the wider context, including computer language tasks (e.g., code translation); (3) evaluating the components of the architecture with respect to helping developers assure robustness and auditors determine compliance of the LLM-based systems.

\begin{ack}
This work was partially supported by financial and other means by the following research projects: DUCA (EU grant agreement 101086308), FLA (supported by the Bavarian Ministry of Economic Affairs, Regional Development and Energy), the DiProLeA (German Federal Ministry of Education and Research, grant 02J19B120 ff), as well as our industrial partners in the FinComp project. We thank the reviewers for their valuable comments.
\end{ack}



\bibliography{refs}

\end{document}